\documentclass[9pt, sigconf, nonacm]{acmart}
\makeatletter
\def\@mkbibcitation{}

\makeatother

\usepackage{amsmath} % For math environments
\usepackage{amsfonts} % For special math fonts like \mathbb
\usepackage{algorithm}
\usepackage[noend]{algorithmic}
\usepackage{xcolor}
\usepackage{booktabs} % For formal tables
\usepackage{graphicx} % Required for including images
\usepackage{array} 
\usepackage{algorithm}
\usepackage{tabularx} % For tables that auto-adjust column width
\usepackage{subcaption}
\usepackage{graphicx}
\usepackage{textcomp}
\usepackage{xcolor}
\usepackage{rotating} % For text rotation
\usepackage{multirow}
\usepackage{tabularx}
\usepackage{diagbox}
\usepackage{caption}
\usepackage{footnote} % For footnotes

\newcolumntype{Y}{>{\centering\arraybackslash}X}

\newcommand{\todo}[1]{\textcolor{black}{#1}}

\newcommand{\ydel}[1]{}

\newcommand{\yadd}[1]{\todo{#1}}

\newcommand{\rqin}[1]{\textcolor{black}{#1}}

\begin{document}

% \title{Unleash the power of Computing-In-Memory in the era of Large Language Models: Robust CiM-backed RAG}
% \title{Towards the potential of Computing-In-Memory in LLMs: Robust CiM-backed RAG}
% \title{Towards the potential of Computing-In-Memory in LLMs: Robust CiM-backed RAG}
\title{Robust Implementation of Retrieval-Augmented Generation on Edge-based Computing-in-Memory Architectures}
% \title{Optimization of Retrieval-Augmented Generation for Robust Implementation on Edge-based Computing-in-Memory Architectures}

\author{Ruiyang Qin$^{1}$, Zheyu Yan$^{1}$, Dewen Zeng$^{1}$, Zhenge Jia$^{1}$, Dancheng Liu$^{2}$, Jianbo Liu$^{1}$, Ahmed Abbasi$^{1}$,Zhi Zheng$^{1}$, Ningyuan Cao$^{1}$, Kai Ni$^{1}$, Jinjun Xiong$^{2}$, Yiyu Shi$^{1}$}

\affiliation{
    \institution{$^{1}$University of Notre Dame $^{2}$University at Buffalo--SUNY}   
    \city{}
    \state{}
    \country{}
}

\begin{abstract}

Large Language Model\yadd{s} (LLM\yadd{s}) deployed on edge devices\ydel{, it typically} learn\ydel{s} through fine-tuning and updating \yadd{a} certain \yadd{portion of their} parameters. Although such learning methods can be optimized to reduce resource utilization, the overall required resources remain a heavy burden on \ydel{the limited resource budget of}edge devices. \yadd{Instead,} Retrieval-Augmented Generation (RAG), a resource-efficient LLM learning method, can improve the \yadd{quality of the} LLM-generated content without updating model parameters. However, the RAG-based LLM may involve repetitive searches on the profile data in every user-LLM interaction. This search can lead to \ydel{increased}\yadd{significant} latency along with the accumulation of user data. \yadd{Conventional} efforts to decrease latency result in restricting the size of saved user data, thus reducing the scalability of RAG as user data continuously grows. 
It remains an open question: how to free RAG from the constraints of latency and scalability on edge devices?
\ydel{We observe that Computing-in-Memory (CiM) architectures can accelerate matrix multiplications and alleviate the usage burden of core computational resources like RAM.}
In this paper, we propose a novel framework to accelerate RAG via \ydel{CiM}\yadd{Computing-in-Memory (CiM) architectures}. 
\yadd{It accelerates matrix multiplications by performing in-situ computation inside the memory while avoiding the expensive data transfer between the computing unit and memory.}
Our framework, \textbf{Ro}bust \textbf{C}iM-backed \textbf{R}AG (RoCR), utilizing \yadd{a novel} contrastive learning\yadd{-based training method}\ydel{, data construction methods,} and noise-aware training, can enable RAG to efficiently\ydel{accomplish} search\ydel{of} profile data with CiM. To the best of our knowledge, this is the\ydel{very} first work utilizing CiM to accelerate RAG.

% In this paper, we propose a novel optimization framework accelerate RAG for Edge LLMs
% In this paper, we propose a robust CiM-backed RAG framework, utilizing contrastive learning and selective write-verify to enhance the noise-resilient capability of data embeddings, tailor data embeddings to be better compatible with CiM computation, and mitigate write errors on semantically similar data embeddings. Considering both algorithm and device levels, our framework performs a co-optimization to enhance the CiM-backed RAG system. Our experiments show that \textcolor{red}{[add some experimental results]}. \textcolor{blue}{the entire abstract needs to be rewrote to correspond with the updated following content}

\end{abstract}

% \keywords{On-device learning, Compute-in-memory (CiM), natural language processing, large language model (LLM)}

\maketitle

\section{Introduction}

The emerging Large Language Models (LLMs) are deployed primarily on centralized cloud platforms \cite{bevilacqua2023automated, qin2024fl} (Cloud LLMs), raising concerns about user privacy and trustworthy issues \cite{neel2023privacy}. These issues become even more prominent in areas such as healthcare \cite{Karabacak_Margetis_2023}, companionship \cite{xu2023large}, and personal assistance \cite{li2024personal}, where the user privacy and trustworthiness of LLMs are crucial. To address these issues, the cloud LLMs will eventually transform into personalized LLMs, capable of generating personalized responses, deployed on edge devices (Edge LLMs), where users can keep all their private data and the model learns from those data locally. 

To better suit the needs of individual users, Edge LLMs must learn from user interactions. However, their capability of learning is constrained by their limited RAM and computational power. Similar to Cloud LLMs, the Edge LLMs primarily learn by fine-tuning their model parameters. Yet, given that these models often contain over 3 billion parameters, updates can be challenging, even with numerous efforts to accelerate them \cite{qin2023enabling, frantar2022gptq, zhu2023pockengine}. For example, using the experimental high-performance embedded system like NVIDIA-AGX, the pockengine method \cite{zhu2023pockengine} can still take 90 hours to learn from a middle-sized dataset Alpaca with only 52k documents, making this option impractical for normal users. 

\begin{figure}[t]
  \centering
  \includegraphics[trim=0 282 360 0, clip, width=1.\linewidth]{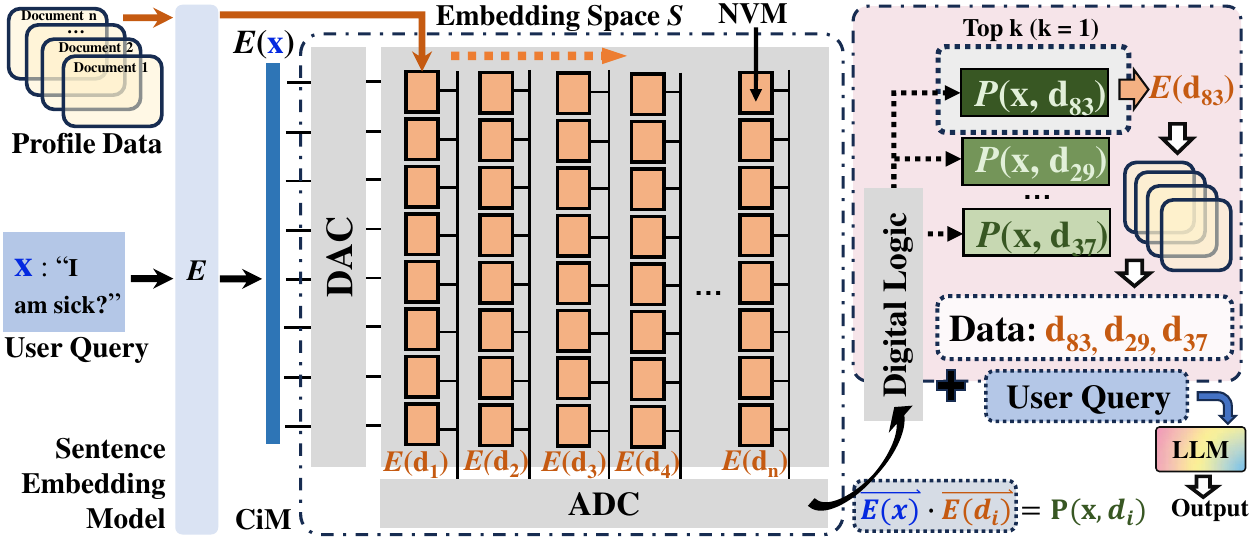}
  \caption{The workflow of RAG on edge-based CiM. CiM \ydel{operates}\yadd{performs} max inner product search (MIPS) to retrieve the top-ranked documents, concatenating them with user query to allow the LLM to generate \ydel{output}personalized \yadd{responses}.}\label{fig:2}
\end{figure}

Retrieval-augmented generation (RAG), on the other hand, is a more resource-efficient choice \cite{lewis2020retrieval}, and hence becoming the de facto learning method for Edge LLMs. In a typical RAG system, it consists of a retriever and a generator. The retriever is commonly backed by max inner product search (MIPS). When the retriever receives a user query, it will retrieve the most relevant document from profile data, as shown in Figure ~\ref{fig:2}. The profile data has many documents, and each document $d_i$ contains specific information that may be relevant to user queries. The generator can be seen as a LLM, which takes the user query $x$ and retriever-obtained documents as a prompt and generates a corresponding response. For every document $d_i$ and the user query $x$, RAG utilizes a sentence embedding model shown in Figure ~\ref{fig:2} to convert them into vectors (\emph{i.e.}, $E(d_i)$ and $E(x)$, respectively). The vectors for documents can be named as document embeddings and stored as a matrix as shown in Figure~\ref{fig:2}. The vector for user query, named query embedding $E(x)$, will be used in MIPS to perform inner product with every document embedding. The larger the product $P(x,d_i)$, the more semantic similar it will be between the user query and the document.

Using RAG, Edge LLMs can provide user-preferred responses by retrieving relevant documents from profile data, and the profile data can be incrementally updated with new documents. This is an efficient learning process without costly updating the model parameters via fine-tuning \cite{hu2021lora}. Other than the inevitable LLM inference cost, the primary computational cost of RAG is about retrieval, which is more than ten times less than the cost of updating model parameters.

While the computational cost of RAG is more edge-friendly, there still exist two issues impeding RAG from being deployed for real-time user interaction on Edge LLMs. Firstly, the growing profile data as stored cannot be unlimited without affecting the access time. \ydel{RAG can keep all user data on RAM and retrieve the appropriate data from RAM instantly. However, RAM is limited and many other programs also need RAM to properly function. While RAG stores and retrieves the user data on} If the size of the profile data exceeds the RAM capacity, it will need to be offloaded into the storage, such as a hard disk drive (HDD) or solid-state drive (SSD). Accessing data from HDD or SSD will significantly increase the data transfer latency~\cite{kang2013enabling}, rendering real-time user interaction impractical. Secondly, the core retrieval method of RAG, MIPS, may experience decreased efficiency as profile data grows, and it can become potentially prohibitive when dealing with overwhelmingly large datasets. For example, on Raspberry Pi 4B, MIPS can take 5 minutes to find one appropriate profile data among 21M documents~\cite{lewis2020retrieval}, which is even longer than the 2-minute inference time of an Edge LLM. Unfortunately, \ydel{there are} few efforts have been made to optimize RAG towards Edge LLMs.

Thus, we propose to utilize the Computing-in-Memory (CiM) architecture to address this issue. As shown in Figure ~\ref{fig:2}, CiM architectures using memory arrays have shown substantial promise in accelerating matrix-vector multiplication \cite{banagozar2019cim}, which \ydel{can be used in}is the key operation of MIPS. The CiM architectures often utilize massive parallel processing to perform computations directly within the memory array where the data is stored, such that they can minimize the data movement through in-situ data access and significantly increase the throughput \cite{sze2017efficient}. Given the same amount of documents, CiM can finish computation within 50ms \cite{peng2019dnn+}, which is negligible compared to the computation latency on normal edge devices. Furthermore, by incorporating non-volatile memory (NVM) devices, such as phase-change memories (PCMs), resistive random-access memories (RRAMs), and ferroelectric field-effect transistors (FeFETs), CiM can outperform conventional MOSFET-based designs in terms of energy efficiency \cite{chen2016eyeriss}.

\begin{figure}[h]
  \centering
  \includegraphics[width=.65\columnwidth]{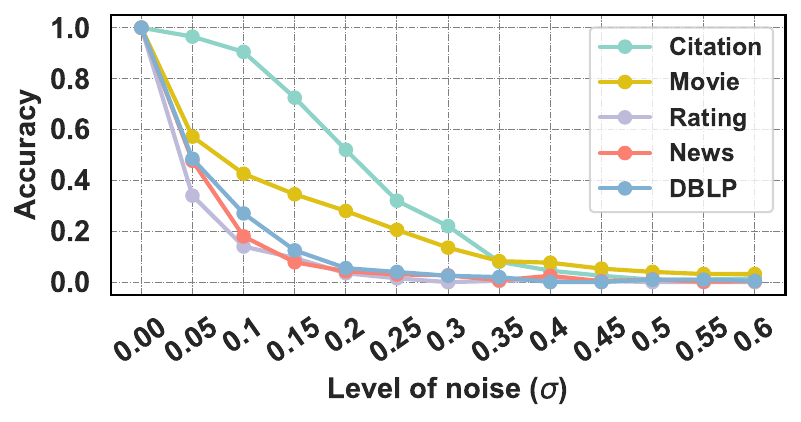}
  % \vspace{-0.2cm}
  \caption{The impact on MIPS accuracy when the RAG's document embedding is \ydel{corrupted with}\yadd{perturbed by} various levels of Gaussian noise caused by the device variations.
  \ydel{The accuracy is the ratio of retrieved documents with and without noise impacts.}
  \yadd{An accurate retrieval means the document retrieved under the impact of the noise is the same as that retrieved without noise.}
  % \vspace{-0.2cm}
  }
%Assuming the MIPS results on the documents without Gaussian noise are correct, accuracy can be calculated by dividing them with the MIPS results on corrupted documents. }
  \label{fig:prelim}
\end{figure}

\ydel{Therefore, the CiM architecture seems to emerge as a promising hardware solution for optimizing RAG. With its parallel processing capabilities and minimal interference with core computational resources dedicated to other edge programs \cite{yang2020co}, leveraging the CiM architecture can effectively diminish RAG latency and enhance its scalability across large-scale data spaces.}

Unfortunately, simply changing the underlying hardware is not enough, as the non-idealities of the NVM devices in CiM array could greatly deteriorate the RAG performance. First, the operations performed in CiM architectures are susceptible to various sources of noise, including electronic noise (thermal, shot, and flicker), device-to-device variability, and line noise from the supporting circuitry \cite{yan2024compute}. These noise sources can corrupt the computations, especially when the signal levels are close to the noise floor, which is a common scenario in high-precision tasks. Such noise issues are critical in RAG applications where the accuracy and quality of the generated content heavily rely on the precision of the underlying computations. Additionally, the CiM architecture is primarily designed and optimized for low-resolution computation \cite{jeong2022variation}. Moreover, CiM arrays are typically sized at a fixed dimension, such as 64x64~\cite{jiang2020device}, which is
different from the documents' embedding dimension (\emph{e.g.}, 128). Therefore, both RAG's data precision (typically FP32) and its embedding dimension need to be
reduced to fit in the size of CiM's crossbar arrays. To illustrate the impact of these on RAG, as an example, we present a preliminary study on MIPS performance in Figure ~\ref{fig:prelim}, where we use \ydel{naive}\yadd{a simple yet representative} Gaussian noise to simulate the noise from \yadd{the device variations in} CiM. As shown in Figure ~\ref{fig:prelim}, as the noise level increases, MIPS accuracy (specified in section 4.1.3) drops dramatically, approaching random guessing. 

% \ydel{We}\yadd{As an example, we simulate the effectiveness of naively using an RRAM-based CiM platform to accelerate the RAG process. The specific setup is discussed in section~\ref{sect:setup}.why} \ydel{use a CiM device RRAM, whose device variation can be found in Table ~\ref{tab:var}, and} \textcolor{red}{[MAY REDRAW THE FIGURE]} examine different levels of device variations, \emph{i.e.}, different $\sigma$ values in the Gaussian noise model shown in equation ~\ref{eq:dev_var}. We evaluate MIPS performance by comparing its retrieved results without device variation to those under device variation. As shown in Figure ~\ref{fig:prelim}, as the noise level increases, MIPS accuracy drops dramatically, approaching random guessing.

To address these issues, we \yadd{further} propose a novel optimization framework for CiM-backed RAG, called \textbf{Ro}bust \textbf{C}iM-backed \textbf{R}AG (RoCR). The framework consists of three parts. The first part is a contrastive learning method. We use it to optimize the document embedding model.
% so that the embedding model can
% generate document and user query embeddings with high noise-resilient capabilities, while such embeddings can fit into CiM architectures with different designs. 
The second part is a novel data construction method to generate both positively and negatively labeled data pairs for contrastive learning. For the profile data, they can be either labeled to indicate the explicit user-preferred response to certain input, or simply statements without explicit labels that only implicitly indicate user preferences. Our data construction method is capable of dealing with both types of profile data. The third part is a noise-aware training method. It goes in tandem with contrastive learning to obtain a sentence embedding model that can generate document and user query embeddings with high noise-resilient capability, while such embeddings can fit into CiM architectures under different designs and configurations.

% To address these issues, we propose a novel optimization framework for CiM-backed RAG, \textbf{Ro}bust \textbf{C}iM-backed \textbf{R}AG (RoCR). The framework consists of three parts. The first part is an end-to-end optimization of RAG to address the noise issues of CiM. For various CiM architectures, their NVM devices may exhibit different ranges of noise and be compatible with specific array sizes, this requires embeddings to be shaped to certain lengths and precisions. Our framework optimizes RAG performance and enables it to adapt to all types of CiM architectures. The second part is a contrastive learning approach to train the embedding model for RAG. User data may either be explicitly labeled by users to indicate preferences or left as purely informational statements without labels specifying user preferences. Contrastive learning is an effective method to enhance the performance of the embedding model, whether using labeled or unlabeled data \textcolor{red}{[cite]}. In contrastive learning, we modify the triplet loss training objective to fit with our designed data augmentation methods. The third part is new data augmentation methods to deal with labeled and unlabeled data respectively. Our method automatically creates triplets that highlight the distinction of every single embedding. By training the contrastive learning model with these constructed data, the robustness of RAG's implementation on CiM architecture is significantly improved. \textcolor{red}{[this need to rewrite to corresponding the following content]}

Our major contributions can be summarized as:
\begin{itemize}
    \item We propose the first work to harvest CiM advantages for RAG acceleration on the edge. We provide a pathway to utilize emerging CiM devices to expand the Edge LLMs' capability in terms of storing a high volume of profile data with fast MIPS computing.
    %\item For the RAG in edge devices, our proposed framework allows it to use CiM to maintain a huge volume of profile data and complete MIPS with minor latency.
    \item We introduce noise-aware training to enhance the noise-resilient capabilities of RAG's document embedding. 
    %Since generating embeddings is costly for the sentence embedding model, 
    The resulting noise-resilient embeddings can be reused robustly, saving resources needed to calibrate and regenerate embeddings.
    \item Our experiments on various datasets show that our proposed framework can improve the RAG performance on multiple CiM devices up to 35\%, approaching to the theoretical RAG performance. Across a wide device variation (noise) range on a single CiM device, our proposed framework can still improve the RAG performance.
\end{itemize}

\section{Related Work}

\subsection{CiM Architectures and their NVMs}

% As shown in Figure~\ref{fig:2}, MIPS can be divided into two steps: (1) vector-matrix multiplication (VMM) between user query embeddings (vector) and user history data embeddings (matrix) and (2) top-K ranking to find the fittest user history data. Step 1 is the most intensive part, so the core building block of nvCiM MIPS accelerators is the crossbar array VMM architecture. 

As shown in the middle part of Figure~\ref{fig:2},
memory arrays are the key component for vector-matrix multiplication. In this array, matrix values are stored at NVM cells, such as emerging NVM technologies like PCMs, RRAMs, and FeFETs, at the cross-points of vertical and horizontal lines. Simultaneously, vector values flow along the horizontal lines of the array. Operations within the memory array take place in the analog domain by exploiting law of physics directly. However, for other essential functions like shift-and-add for multiple bits and sorting to find the top-k ranked values would be done in the digital domain. Thus, digital-to-analog and analog-to-digital converters (DACs and ADCs) are used to connect these different components.

CiM arrays suffer from various sources of variations and noises. Two major ones include spatial variations and temporal variations. Spatial variations result from fabrication defects and have both local and global correlations. FeFET devices also suffer from temporal variations due to the stochasticity in memory switching and also aging, which causes fluctuations in conductance when programmed at different times. Temporal variations are typically independent from device to device and are irrelevant to the value to be programmed~\cite{feinberg2018making}.   
In this work, as a proof of concept, we focus on the impact of temporal variations in the programming process on DNN performance. Temporal variation makes the programmed resistance of a device deviate from what is expected. The proposed framework can also be extended to other sources of variations with modification.

Measurement results~\cite{yan2022swim, yan2021uncertainty} show that the noise on DNN weights caused by device variations can be safely modeled as a Gaussian noise with zero mean, each with a standard deviation associated with the weight value. A detailed representation is given by:
\begin{equation}\label{eq:dev_var}
    \mathbf{v} = \mathbf{v_0} + \Delta\mathbf{v}, \Delta\mathbf{v}\sim\mathcal{N}(0,\sigma_v)
\end{equation}
where $\mathbf{v}$ is the actual embedding deployed on the accelerators, $\mathbf{v_0}$ is the target embedding value, and $\sigma_v$ is a value measured by the experiments. We collect the measurement results from RRAM and FeFET devices and the specific value will be discussed in Section~\ref{sect:setup}.

\begin{figure}[t]
  \centering
  \includegraphics[trim=0 445 748 0, clip, width=1.\linewidth]{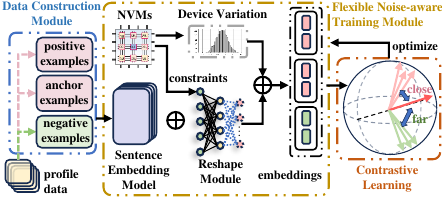}
  \caption{Overview of \ydel{our}\yadd{the proposed} Robust CiM-backed RAG framework (RoCR). It \ydel{can} optimize\yadd{s} the sentence embedding model to adapt different types of NVMs utilized by CiM.}
  \label{fig:framework}
\end{figure}

\subsection{Past Noise Mitigation Methods}

Several strategies have been introduced to tackle the challenge of device variations in CiM accelerators. These methods can be separated into software and hardware-based techniques.

The software-based techniques are generally developed to obtain more robust DNN models~\cite{jiang2020device, yan2021uncertainty, gao2021bayesian, yan2023improving} or recommendation systems~\cite{li2022imars}, and are thus not suitable for generating more robust MIPS solutions.

For the hardware techniques, the write-verify procedure \cite{shim2020two, yao2020fully} is one of the most commonly used approach during programming. Initially, a NVM device is programmed to a set state via a designated pulse pattern. Subsequent to this, the device's value is verified to ascertain if its conductance aligns with a stipulated range of the desired value, essentially assessing its accuracy. If discrepancies arise, a supplemental update pulse is initiated to reset the device conductance nearer to the target. This loop persists until the disparity between the programmed device value and the target value diminishes to a satisfactory margin, typically taking a handful of cycles. Cutting-edge research suggests that by selectively applying write-verify to a subset of pivotal devices, one can uphold the average accuracy of a DNN \cite{yan2022swim}. Additionally, a variety of circuit design initiatives~\cite{shin2021fault, jeong2022variation} have been put forth to counteract device variations.

\section{Proposed work}

% \textcolor{red}{[we drop the intro part here due to space...]}
% In this section, we first provide an overview of our framework. Then, we explain details of each piece of the framework, starting from the mechanism of operating MIPS on CiM and the impact of MIPS results due to the device variation. After that, we demonstrate the flexible noise-aware training method to improve the sentence embedding model capabilities of generating noise-resilient embedding. To enable our training method, we offer two data construction methods. Additionally, we explain how our framework improve the robustness of CiM-backed RAG.  
% \textcolor{red}{[can delete it depends on space]}

\subsection{Framework Overview}

As shown in Figure ~\ref{fig:framework}, our proposed framework, \textbf{Ro}bust \textbf{C}iM-backed \textbf{R}AG (RoCR), consists of three stages. First, we apply contrastive learning to utilize the training data to optimize the training module. To do that, in the second stage, we take the  profile data and construct via a data construction module to obtain contrastive training data pairs, which are then used in the flexible noise-aware training module. In the third stage, we obtain the constraints of NVMs in CiM via profiling. These constraints will be encoded into the flexible noise-aware training module and used to train the sentence embedding model so that it can  generate embedding that are robust against device variation of the target NVMs. After training, the training module can be turned into a new sentence embedding model and generate CiM-friendly embeddings.

\subsection{Contrastive Learning: Triplet Loss Function} 
% \textcolor{orange}{remove the figure 1 references here, or discuss them in introduction.  Jenny: Agree, this has been very far away from the graph...}

When we apply RAG using CiM, we first need to store embeddings into NVMs as shown in Figure ~\ref{fig:2}. Such embeddings are generated by the sentence embedding model, and they are the numerical representations of profile data. Each single document in the profile data can have its unique embedding, which is a vector. The embeddings stored on NVMs can consist of a matrix as the orange blocks shown in Figure ~\ref{fig:2}. Given a user query, which will also be converted into an embedding, CiM can operate MIPS between this user query embedding and all profile embeddings simultaneously via vector-matrix multiplication. The top-ranked values in the product will be used as the index to retrieve the corresponding document data, as the pink block shown in Figure ~\ref{fig:2}. This retrieved user-relevant document is the output of MIPS.

\begin{figure}[h]
  \centering
  \includegraphics[trim=0 348 585 0, clip, width=1.\linewidth]{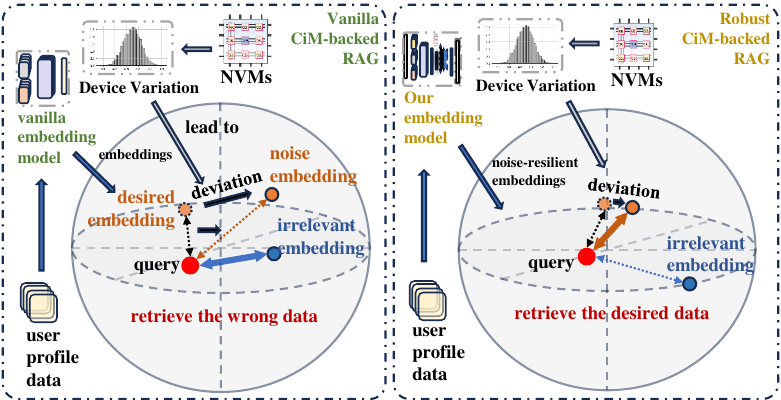}
  \caption{Improvement by our Robust CiM-backed RAG. Our framework generates noise-resilient embeddings, as shown the orange and blue point in right subfigure}
  \label{fig:optimize}
\end{figure}

However, as we have explained in Section 2.1, writing the document embeddings into NVMs can cause them to suffer from temporal variations (device variations). Then, the NVM-stored embeddings will be different from the original sentence embedding model generated embeddings. As shown in Figure ~\ref{fig:optimize}, the vanilla embedding model generates desired embedding, which will deviate to the noise embedding under device variation, such that the irrelevant embedding is ranked higher than desired embedding due to its larger inner product.

Contrastive learning can learn the representations via push away dissimilar examples and pull close similar examples \cite{chen2020simple}. In particular, the contrastive loss function can be used to increase the distance between dissimilar examples. 

In our work, we propose to improve the noise-resilient capability by contrastive learning. By increasing the distance between dissimilar examples, as shown the right subfigure in Figure ~\ref{fig:optimize}, deviated desired embedding will still have a larger inner product with the query compared to the irrelevant embedding. Our contrastive learning loss function is based on Weinberger et al. \cite{weinberger2005distance}. 
For each example \(x_i\) in a mini-batch of N anchor examples, our data construction method will construct \(K\) positive and \(K\) negative examples corresponding to \(x_i\).
% \textcolor{orange}{Is it that: for each sample $x_i$ in a mini-batch of N anchor examples,our data construction method will constrict K positive and K negative examples corresponding to $x_i$?}
% For a set of N anchor examples \(\{x_i\}_{i=1, ..., N}\), the corresponding mini-batch consists of $K$N example combos after our data construction methods, where \(K\) is the number of negative and positive examples to make given one anchor example. 
We can have \(\{\{(x_i, x_i^-, x_i^+)_k\}_{i=1, ..., N}\}_{k=1, ..., K}\), in which $x^-$ and $x^+$ are negative and positive examples corresponding to $x_i$, where $x_i$ is closer to $x_i^+$ compared to $x_i^-$. Also, \(emb(x_i)\) represents the learned embedding of \(x_i\). Then the loss function \(\mathcal{L}\) can be defined as: 
%\jx{please check the following equaiton, $x_i^s$ and $x_i^k$ etc are not definied, $k$ was defined as a subscript in the triplets before.}

\begin{equation}
    \begin{gathered}
        \mathcal{L} = \sum_{i=1}^{N} \frac{1}{K} \sum_{k=1}^{K} \max \left(0, \text{d}(x_i, x_{i(k)}^{-}) - \text{d}(x_i, x_{i(k)}^{+}) + m\right), \\
        \text{d}(x_a, x_b) = \text{sim}(\text{emb}(x_a), \text{emb}(x_b))
    \end{gathered}
\end{equation}
% \begin{equation}
%     \begin{gathered}
%         \mathcal{L} = \sum_{i=1}^{N} \frac{1}{k} \sum_{s=1}^{k} \max \left(0, \text{d}(x_i, x_i^{-k}) - \text{d}(x_i, x_i^{+k}) + m\right), \\
%         \text{d}(x_a, x_b) = \text{sim}(\text{emb}(x_a), \text{emb}(x_b))
%     \end{gathered}
% \end{equation}
%\jx{please check: the following sentence is a repetition of previous ones before the equanation.}
%where $x_i$ is an anchor example, with its corresponding \(K\) negative examples $x_{i(k)}^{-}$ and positive examples $x_{i(k)}^{+}$, where $x^i$ is closer to $x_{i(k)}^{+}$ compared to $x_{i(k)}^{-}$. 
The distance \(d(x_a, x_b)\) is calculated by the Euclidean distance between embeddings of two data \(emb(x_a)\) and \(emb(x_b)\). 
% They formalize a set of triple examples \(\{(x_i, x^-_i, x^+_i)\}_{i=1}^k \). 
The function \(sim()\) calculate the semantic similarity.

\subsection{Data Construction}

To train the sentence embedding model via contrastive learning, it is critical to construct pairs of examples where the positive examples and negative examples need to be distinct from each other \cite{zeng2021positional}. In our work, since we use triplet contrastive loss, instead of pairs of examples, we will construct trios of examples where each triplet contains an anchor, positive, and negative example. 

\begin{figure}[h]
  \centering
  \includegraphics[trim=0 314 660 0, clip, width=1.\linewidth]{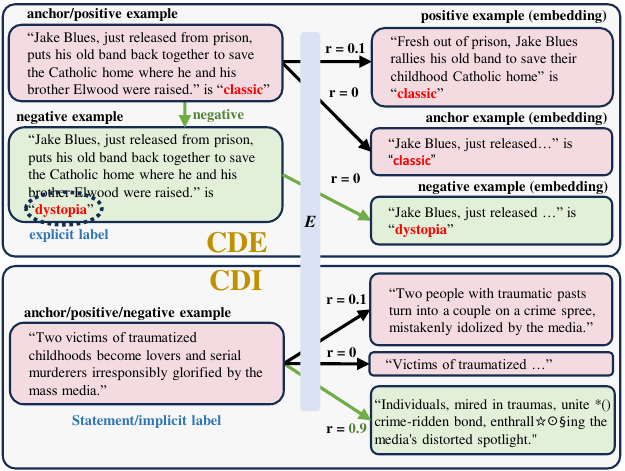}
  % \caption{Construct from data with explicit labels ()}
  \caption{\yadd{Examples of the} two data construction methods. For data with explicit labels, CDE is used to construct the training data. For data without explicit labels (implicit labeled data), CDI is used to construct the training data.}
  % \caption{Example data augmentation method for data with explicit and implicit label \textcolor{red}{[need to explain what is implicit data what is explicit data, use the example in this figure]}}
  \label{fig:ex_im_label}
\end{figure}

We use profile data to construct triplets of examples. For the profile data, it is generated by the user during the user-LLM interaction and contains the user preference information. There exists two situations for such data. First, the profile data can contain explicit labels indicating the user preferred response to the corresponding content. Second, the profile data also can be statements containing the user-related information 
but without explicit user preferences
% \jx{but without explicit user preferences}. 
As shown in Figure ~\ref{fig:ex_im_label}, to deal with the two situations, we come up with two data construction methods: \textbf{C}onstruction \textbf{D}ata with \textbf{E}xplicit labels (CDE) and \textbf{C}onstruction \textbf{D}ata with \textbf{I}mplicit labels (CDI).

\subsubsection{\textbf{\underline{C}onstruction Trios via \underline{D}ata with \underline{E}xplicit Labels (CDE)}}
For the data with explicit labels, each of the data consists of a textual content \textit{c} and its corresponding label \textit{l} which indicates the user preferred response regarding to the content \textit{c}. As shown in the CDE part in Figure ~\ref{fig:ex_im_label}, there exists explicit label circled by dashed line. Using the profile data, we will construct triplet examples in the format of \((x_i, x^-_i, x^+_i)\). Given a dataset \(\mathcal{D}\) with size of $n$ profile documents, each piece of data consists of a content \(c_i\) and the corresponding label \(l_i\) where \(i \in \{1, 2, ..., n\}\). The anchor example \(x_i\) can be constructed as: 
\begin{equation}
    x_i = c_i \oplus l_i, \quad \text{for } i = 1, 2, \ldots, n
\end{equation}
where \(\oplus\) denotes a concatenation operation, specifically used here to combine label and content. 
%For each anchor data \(x_i\),
%with content \(c_i\), 
Negative examples \(x^-_{i}\) can be constructed by concatenating \(c_i\) with a random label 
% \(l_k\) 
% \textcolor{red}{$l_j$ that is different from $l_i$}
$l_j$ that is different from $l_i$
as follows:
%whose corresponding content \(c_k \neq c_i\). \(k \in K\) where \(K\) indicates the number of random sampling and \(i \neq k\). The negative example \(x^-_{i(k)}\) \jx{why subscript $i(k)$? maybe just subscript $i$ is enough?} can be constructed as:
% \begin{equation}
% x^-_{i(k)} = c_i \oplus l_k, \quad \text{for } k = 1, 2, \ldots, K
% \end{equation}

%\begin{equation}
%    x^-_{i} = c_i \oplus l_k, \quad %\text{where } c_i \neq c_k, l_i \neq l_k, 
%\end{equation}
\begin{equation}
    x^-_{i} = c_i \oplus l_j, \quad \text{where } l_i \neq l_j. 
\end{equation}
\noindent
% \jx{where \(k \in K\) with \(K\) indicating the number of random sampling and \(i \neq k\).}
% where \(k \in K\) with \(K\) indicating the number of random sampling and \(i \neq k\).
% \textcolor{orange}{I think the negative equation is malformed. It is not consistent with the data size. It should be $x^-_{i} = c_i \oplus l_k, \quad \text{where } c_i \neq c_k, l_i \neq l_k$}
Randomly assigning a different label ensures diversity in the negative examples while maintaining the same content from the anchor.

Different from constructing anchor and its negative examples, it is challenging to construct positive examples corresponding to the anchor examples since it is more difficult to formalize semantically similar data than to formalize semantically dissimilar data. To construct positive examples, we follow the SimCSE method \cite{gao2021simcse} to add a dropout rate \(r\) into the sentence embedding model \(\mathcal{M}\). The process for constructing positive examples involves two main steps.

First, the textual positive example is formalized as:
% \textcolor{orange}{i=1...n?}
\begin{equation}
    x^+_i = x_i, \quad \text{for } i = 1, 2, ..., n
    \label{eq:pos_label}
\end{equation}
where we align each anchor with the corresponding positive example. This step effectively duplicates the anchor data as a starting point for generating the embeddings.

Second, the embedding generation process varies based on the dropout rate applied within the model \(\mathcal{M}\). When model \(\mathcal{M}\) is utilized to generate embeddings for anchor and negative examples, the dropout rate is set to \(0\). In contrast, for generating embeddings for positive examples, a non-zero dropout rate \(r\) is used. The anchor, negative, positive examples, as shown in Figure ~\ref{fig:ex_im_label}, can be constructed as: 
\begin{equation}
    \begin{aligned}
        emb(x_i) &= \mathcal{M}(x_i, dropout = 0) \\
        emb(x_i^-) &= \mathcal{M}(x_i^-, dropout = 0) \\
        emb(x^+_i) &= \mathcal{M}(x^+_i, dropout = r)
    \end{aligned}
\end{equation}
% \(emb(x_i) = \mathcal{M}(x_i, dropout = 0)\), \(emb(x_i^-) = \mathcal{M}(x_i^-, dropout = 0)\), and \(emb(x^+_i) = \mathcal{M}(x^+_i, dropout = r)\). 
The condition of \(r \neq 0\) can induce variation in the embeddings, enhancing the model's ability to recognize semantically similar yet variably expressed content.

Given the construction factor \(K\), %\jx{shouldn't this be $K$ instead of $k$? same for the equation below}, 
we can construct the triplet data examples as:

\begin{equation}
    \mathcal{D}_{triplet} = \bigcup_{i=1}^N \left\{ (x_{i(k)}, x^-_{i(k)}, x^+_{i(k)}) : k = 1, 2, \ldots, K \right\}
\end{equation}

For the triplet data examples \(\mathcal{D}_{triplet}\), their embeddings for each augmentation \(k\) are given by:

\begin{equation}
    \mathcal{E} = \bigcup_{i=1}^N \left\{ (emb(x_{i(k)}), emb(x^-_{i(k)}), emb(x^+_{i(k)}) : k = 1, 2, \ldots, K \right\}
\end{equation}

As shown in Figure ~\ref{fig:ex_im_label}, for data with explicit labels, a content \(c\) can concatenate with its corresponding label \(l\) to formalize the positive and anchor example. That content \(c\) can also concatenate with other labels \(l'\) to formalize the negative example. The positive example can be finally obtained from the sentence embedding model with dropout rate \(r\). The anchor and negative example can be finally obtained from the sentnece embedding model with \(r = 0\). 

\newcolumntype{Y}{>{\centering\arraybackslash}X}
\newcolumntype{C}{>{\centering\arraybackslash}p{5mm}}  % Fixed width column, adjust the 5mm as needed

\begin{table*}[htbp]
\caption{Performance comparison between our framework and four baselines on five CiM devices with device variation \ydel{\(\sigma = 0.1\)}\yadd{specified in Table~\ref{tab:var}} across five datasets. Evaluate the performance of our framework using EDC (RoCR-EDC) and using IDC (RoCR-IDC) to optimize the performance of RAG, which utilizes Gemma-2 as its LLM.}
% \caption{Gemma-2B performances on five datasets and five CiM devices given the optimization of our framework and four baselines \textcolor{red}{explain what is ex, im. Add arrow to explain metric. explain baseline names}}
\footnotesize
\begin{tabularx}{\textwidth}{Cc|c*{9}{Y}} 
\toprule
\toprule
 \multicolumn{2}{c|}{\textbf{Dataset}} & \multicolumn{2}{c}{\textbf{Citation}} & \multicolumn{2}{c}{\textbf{Movie}} & \multicolumn{2}{c}{\textbf{Rating}} & \multicolumn{2}{c}{\textbf{News}} & \multicolumn{2}{c}{\textbf{DBLP}} \\
\midrule
\midrule
CiM& Method & Acc \(\uparrow\) & F1 \(\uparrow\) & Acc \(\uparrow\) & F1 \(\uparrow\) & MAE \(\downarrow\) & RMSE \(\downarrow\) & ROUGE-1 \(\uparrow\) & ROUGE-L \(\uparrow\) & ROUGE-1 \(\uparrow\) & ROUGE-L \(\uparrow\) \\
\midrule
\multirow{6}{*}{\rotatebox[origin=c]{90}{Device-1}} 
& SWV & 0.4208 & 0.3339 & 0.1305 & 0.1974 & 0.3850 & 0.8093 & 0.0754 & 0.0731 & 0.1709 & 0.1590 \\
& CxDNN & 0.4223 & 0.3576 & 0.1516 & 0.1762 & 0.4404 & 0.9135 & 0.0640 & 0.0632 & 0.1646 & 0.1449 \\
& CorrectNet & 0.4155 & 0.3791 & 0.0996 & 0.1305 & 0.3609 & 0.7071 & 0.0512 & 0.0764 & 0.1603 & 0.1538 \\
& Vanilla RAG & 0.4401 & 0.3476 & 0.1017 & 0.0838 & 0.3903 & 0.8944 & 0.0754 & 0.0731 & 0.1731 & 0.1473 \\
& \textbf{RoCR-CDE} & \textbf{0.5536} & \textbf{0.3956} & \textbf{0.2242} & \textbf{0.2303} & \textbf{0.3108} & \textbf{0.6856} & \textbf{0.1041} & \textbf{0.0987} & \textbf{0.2066} & \textbf{0.1924} \\
& \textbf{RoCR-CDI} & \textbf{0.5409} & \textbf{0.5117} & \textbf{0.2273} & \textbf{0.2487} & \textbf{0.2767} & \textbf{0.6083} & \textbf{0.0831} & \textbf{0.0808} & \textbf{0.2317} & \textbf{0.2176} \\

\midrule
\multirow{6}{*}{\rotatebox[origin=c]{90}{Device-2}} 
& SWV & 0.1831 & 0.1552 & 0.1992 & 0.1957 & 0.4205 & 0.8775 & 0.0296 & 0.0289 & 0.1968 & 0.1874 \\
& CxDNN & 0.4013 & 0.3557 & 0.2167 & 0.2019 & 0.4423 & 0.8367 & 0.0604 & 0.0791 & 0.1517 & 0.1401 \\
& CorrectNet & 0.3827 & 0.3209 & 0.1625 & 0.1909 & 0.3762 & 0.8062 & 0.0513 & 0.0505 & 0.2042 & 0.1945 \\
& Vanilla RAG  & 0.4801 & 0.3462 & 0.1576 & 0.2079 & 0.4153 & 0.9354 & 0.0296 & 0.0289 & 0.1618 & 0.1353 \\
& \textbf{RoCR-CDE} & \textbf{0.5407} & \textbf{0.4396} & \textbf{0.2924} & \textbf{0.2509} & \textbf{0.2553} & \textbf{0.5385} & \textbf{0.1209} & \textbf{0.0946} & 0.2025 & 0.1906 \\
& \textbf{RoCR-CDI} & \textbf{0.5299} & \textbf{0.4591} & \textbf{0.2971} & \textbf{0.2386} & \textbf{0.2124} & \textbf{0.5763} & \textbf{0.0884} & \textbf{0.0853} & \textbf{0.2240} & \textbf{0.2098} \\

\midrule
\multirow{6}{*}{\rotatebox[origin=c]{90}{Device-3}} 
& SWV & 0.2450 & 0.2564 & 0.1695 & 0.1641 & 0.3460 & 0.7416 & 0.0725 & 0.069 & 0.1018 & 0.0954\\
& CxDNN & 0.4811 & 0.4006 & 0.2367 & 0.2113 & 0.2851 & 0.6928 & 0.0761 & 0.0707 & 0.1425 & 0.1111 \\
& CorrectNet & 0.4510 & 0.3918 & 0.0792 & 0.1029 & 0.3704 & 0.7937 & 0.0585 & 0.0555 & 0.1715 & 0.1346 \\
& Vanilla RAG & 0.4852 & 0.3618 & 0.1614 & 0.1636 & 0.3255 & 0.7649 & 0.0725 & 0.0690 & 0.1647 & 0.1437 \\
& \textbf{RoCR-CDE} & \textbf{0.5139} & \textbf{0.4116} & \textbf{0.2242} & \textbf{0.2215} & \textbf{0.3208} & \textbf{0.6481} & \textbf{0.0825} & \textbf{0.0805} & \textbf{0.1893} & \textbf{0.1754} \\
& \textbf{RoCR-CDI} & \textbf{0.5515} & \textbf{0.4984} & \textbf{0.2152} & \textbf{0.2131} & \textbf{0.2916} & \textbf{0.6245} & \textbf{0.1099} & \textbf{0.1049} & \textbf{0.2294} & \textbf{0.2140} \\

\midrule
\multirow{6}{*}{\rotatebox[origin=c]{90}{Device-4}}  
& SWV & 0.5135 & 0.4260 & 0.1271 & 0.1178 & 0.3610 & 0.8196 & 0.0259 & 0.0256 & 0.1871 & 0.1786 \\
& CxDNN & 0.4733 & 0.3964 & 0.1267 & 0.2158 & 0.3468 & 0.7616 & 0.0646 & 0.0634 & 0.1603 & 0.1538 \\
& CorrectNet & 0.4628 & 0.4019 & 0.1592 & 0.1847 & 0.4013 & 0.9274 & 0.0705 & 0.0750 & 0.1628 & 0.1292\\
& Vanilla RAG & 0.2101 & 0.2401 & 0.1219 & 0.2019 & 0.4015 & 0.8544 & 0.0505 & 0.0489 & 0.1929 & 0.1814\\
& \textbf{RoCR-CDE} & \textbf{0.5836} & \textbf{0.5555} & \textbf{0.1706} & \textbf{0.2817} & \textbf{0.3139} & \textbf{0.6856} & \textbf{0.0873} & \textbf{0.0851} & \textbf{0.1984} & \textbf{0.1882} \\
& \textbf{RoCR-CDI} & \textbf{0.5352} & \textbf{0.4289} & \textbf{0.1642} & \textbf{0.2445} & \textbf{0.2706} & \textbf{0.5916} & \textbf{0.1154} & \textbf{0.1128} & \textbf{0.2148} & \textbf{0.1978} \\

\midrule
\multirow{6}{*}{\rotatebox[origin=c]{90}{Device-5}} 
& SWV & 0.4320 & 0.3541 & 0.1250 & 0.1076 & 0.3652 & 0.7616 & 0.0434 & 0.0427 & 0.0985 & 0.0923\\
& CxDNN & 0.4301 & 0.0538 & 0.0751 & 0.0458 & 0.3503 & 0.8185 & 0.0707 & 0.0682 & 0.2042 & 0.1945 \\
& CorrectNet & 0.4145 & 0.3926 & 0.1083 & 0.1395 & 0.5526 & 0.8185 & 0.0735 & 0.0776 & 0.2096 & 0.1879\\
& Vanilla RAG & 0.4256 & 0.3522 & 0.0847 & 0.0863 & 0.3951 & 0.8515 & 0.0676 & 0.0653 & 0.2018 & 0.1846 \\
& \textbf{RoCR-CDE} & \textbf{0.5698} & \textbf{0.5223} & \textbf{0.2152} & \textbf{0.1669} & \textbf{0.2959} & \textbf{0.6245} & \textbf{0.0936} & \textbf{0.0891} & 0.1946 & 0.1844 \\
& \textbf{RoCR-CDI} & \textbf{0.5254} & \textbf{0.4504} & \textbf{0.2394} & \textbf{0.2458} & \textbf{0.2624} & \textbf{0.6325} & \textbf{0.0799} & \textbf{0.0764} & \textbf{0.2238} & \textbf{0.2095} \\

\bottomrule
\bottomrule
\end{tabularx}
\label{tab:main}
\end{table*}

\subsubsection{\textbf{\underline{C}onstruction Trios via \underline{D}ata with \underline{I}mplicit Labels (CDI)}} For data with implicit labels, each of the data consists of solely textual content \textit{c}.
%contains the user related information. 
As shown of the CDI part in Figure ~\ref{fig:ex_im_label}, there is no explicit label to indicate 
user preferences.
%what user specifically expect. 
Instead, the data can be seen as a statement containing some user-related information. To construct the anchor examples and positive examples, we can use the exact same method in EDC. Given a dataset \(\mathcal{D}\) with size of \textit{n} profile data, each piece of data consits of a content \(c_i\). The anchor data \(x_i\) can be constructed as:
\begin{equation}
    x_i = c_i, \quad \text{for } i = 1, 2, \ldots, n
\end{equation}

For each anchor data \(x_i\), constructing its corresponding negative example is not as simple as merely concatenating the content \(c_i\) with a non-corresponding label \(l_k\). To construct negative examples, we employ a reciprocal approach with the positive examples, applying a similar method to both.

We first initialize the negative example and positive example following the equation ~\ref{eq:pos_label}:
\begin{equation}
    x_i^- = x_i^+ = x_i, \quad \text{for } i = 1, 2, \ldots, n
    \label{eq:pos_anc_neg}
\end{equation}

For the positive example \(x_i^+\), it can be finalized by incorporating a dropout rate \(r\) into the sentence embedding model \(\mathcal{M}\), where a rate of \(0 < r \leq 0.2\) can generate a sentence embedding with a semantic representation similar to \(x_i\) and ensure good model training performance \cite{gao2021simcse}. Increasing the dropout rate to a higher value, such as 0.5, can distort the semantic representation of \(x_i^+\), making it dissimilar to that of \(x_i\). Training the model with such positive examples can result in poorer performance. For positive examples in training the sentence embedding model, the higher dropout rate performs more like a noise rather than a data augmentation method.

In our work, we train the sentence embedding model to generate embeddings that maintain their integrity under noisy conditions, such as during writing into Compute-in-Memory (CiM). The noise can alter or fragment the original semantic representations. For instance, as illustrated in Figure \ref{fig:ex_im_label}, using a high dropout rate \(r = 0.9\) can lead to a negative example with a corrupted representation. Although it may lack certain informative content, this negative example becomes semantically distinct from both the anchor and positive examples, effectively simulating the effect of CiM corruption. This approach not only differentiates the negative examples semantically but also aligns them with the corrupted data scenarios for noise-aware training.

Given the triple examples \((x_i, x_i^-, x_i^+)\), for \(i = 1, 2, ..., n\) as shown in equation ~\ref{eq:pos_anc_neg}, we have the dropout rate \(r\) for formalizing the positive examples where \(0 < r \leq 0.2\). Correspondingly, the dropout rate for formailzing the negative examples can be \(1 - r\). Given the sentence embedding model \(\mathcal{M}\), the anchor example, positive example, and negative example can be constructed as: 
\begin{equation}
    \begin{aligned}
        \text{emb}(x_i) &= \mathcal{M}(x_i, \text{dropout} = 0) \\
        \text{emb}(x_i^-) &= \mathcal{M}(x_i^-, \text{dropout} = 1 - r) \\
        \text{emb}(x^+_i) &= \mathcal{M}(x^+_i, \text{dropout} = r)
    \end{aligned}
\end{equation}
% \(\text{emb}(x_i) = \mathcal{M}(x_i, \text{dropout} = 0)\), \(\text{emb}(x_i^-) = \mathcal{M}(x_i^-, \text{dropout} = 1 - r)\), and \(\text{emb}(x^+_i) = \mathcal{M}(x^+_i, \text{dropout} = r)\).

\subsection{Flexible Noise-aware Training}

In the previous two stages, we construct the data to train the sentence embedding model based on contrastive learning. Meanwhile, the training can be more effective when injecting the simulated device variation \cite{zur2009noise} so that the model can be optimized with consideration of the device variation. Additionally, the sentence embedding model needs to produce embeddings that can fit with the different CiMs, which might have various NVM designs. To do that, we need the sentence embedding model reshapes its output embeddings into certain dimensions and precision. Hence, we propose a flexible noise-aware training method, which can generate the noise-resilient embedding, fitting to various CiMs.

As shown in Figure ~\ref{fig:framework}, in the flexible noise-aware training module, the embedding generated by sentence embedding model will be shaped based on the CiM's NVMs constraints where required dimension is \(d\) and required precision is \(p\), and being injected device variation to formalize the embeddings. The reshape module, shown in Figure ~\ref{fig:framework}, seen as an autoencoder to reconstruct its input embedding \cite{malireddy2020scar}, can be expressed as \(shp()\), initialized by \(d\) and \(p\), takes the anchor embedding \(emb(x_i)\) as input. We can have \(shp(emb(x_i)) = emb(x_i)^{d*p}\). Based on the device variation shown as Table ~\ref{tab:var}, we can have:
\begin{equation}
    emb(x_i)_{\sigma}^{d*p} = (e' * L_0 + e' * L_1 + e' * L_2 + e' * L_3) * \sigma,
\end{equation}
% \(emb(x_i)_{\sigma}^{d*p} = (e' * L_0 + e' * L_1 + e' * L_2 + e' * L_3) * \sigma\), 
where \(e' = emb(x_i)^{d*p}\). The device variation, as noise, is injected into embeddings to formalize \(emb(x_i)_{\sigma}^{d*p}\), which will be used in contrastive learning to train the sentence embedding model, as shown in Figure ~\ref{fig:framework}.

% \textcolor{blue}{In our work, we propose a flexible noise-aware contrastive learning method to improve the sentence embedding noise-resilient capability in various shapes made by compression and quantization. We construct training data into three groups including anchor, positive, and negative. While our method pull anchor and positive close and push anchor and negative further during the training, it can inject noise into the generating embeddings to calibrate the model to adapt the noise and eventually learn to generate noise-resilient embedding.}

% \input{tables/table_1}

\section{Experimental Evaluation}

\subsection{Experimental Setup}\label{sect:setup}

\subsubsection{\textbf{Datasets.}} To demonstrate our robust CiM-backed RAG, we employ five datasets with different tasks and domains, including Citation Identification \cite{salemi2023lamp} (\textbf{Citation}), Movie Tagging \cite{harper2015movielens} (\textbf{Movie}), Product Rating \cite{ni2019justifying} (\textbf{Rationg}), News Headline Generation \cite{misra2022news} (\textbf{News}), and DBLP-Citation-network V14 \cite{tang2008arnetminer} (\textbf{DBLP}) to evaluate the proposed framework. The data in each dataset consists of query data and profile data. In our evaluation, the profile data will be used to formalize user history, and the profile corresponding query data will be used as the user input. The first three datasets contain binary, five-class, and fifteen-class classification tasks respectively. The last two datasets contain text generation tasks. In the Citation Identification dataset, every piece of query data consists of a paper title and two references, and the correct reference is provided. RAG uses the profile data corresponding to the paper titles with their detailed contents to choose the appropriate reference. In the Movie Tagging dataset, each query data contains a description of a movie, and RAG uses a similar description and its corresponding tag in the profile data to tag the query data. The Product Rating dataset has a similar structure as the Movie Tagging dataset. In News Headline Generation and DBLP datasets, each query data contains an abstract, which can be summarized into a title. RAG uses a similar abstract and its corresponding title in profile data to generate the title for query data. All five datasets have labels in their query data.

\begin{figure*}[ht]
  \centering
  % First row of figures
  \begin{subfigure}[b]{0.237\textwidth}
    \includegraphics[width=1\textwidth]{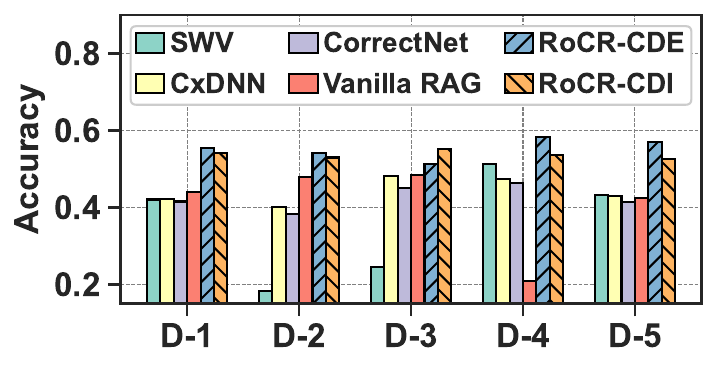}
    \caption{Citation on Gemma-2B}
    \label{fig:a}
  \end{subfigure}
  % \hfill % space between the subfigures
  \begin{subfigure}[b]{0.237\textwidth}
    \includegraphics[width=1\textwidth]{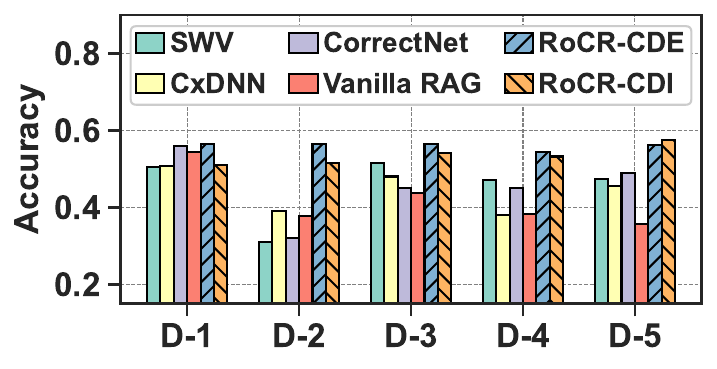}
    \caption{Citation on Phi-2}
    \label{fig:b}
  \end{subfigure}
  % \hfill % space between the subfigures
  \begin{subfigure}[b]{0.237\textwidth}
    \includegraphics[width=1\textwidth]{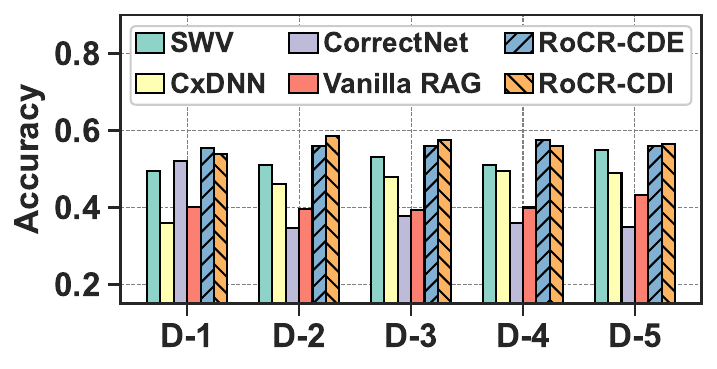}
    \caption{Citation on Mistral-7B}
    \label{fig:c}
  \end{subfigure}
  % Second row of figures
  \begin{subfigure}[b]{0.237\textwidth}
    \includegraphics[width=1\textwidth]{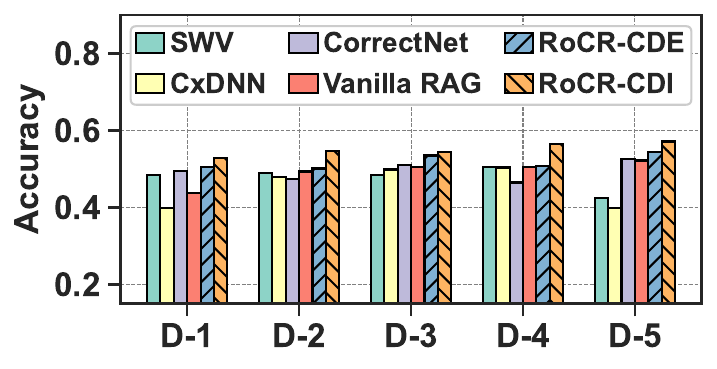}
    \caption{Citation on Llama-2-3B}
    \label{fig:d}
  \end{subfigure}
  % \hfill % space between the subfigures
  \begin{subfigure}[b]{0.237\textwidth}
    \includegraphics[width=1\textwidth]{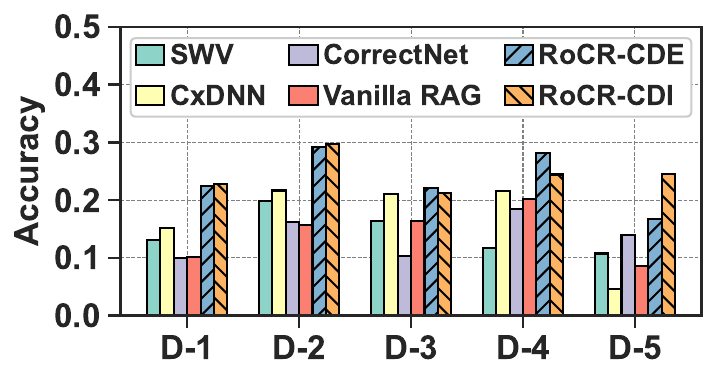}
    \caption{Movie on Gemma-2B}
    \label{fig:e}
  \end{subfigure}
  % \hfill % space between the subfigures
  \begin{subfigure}[b]{0.237\textwidth}
    \includegraphics[width=1\textwidth]{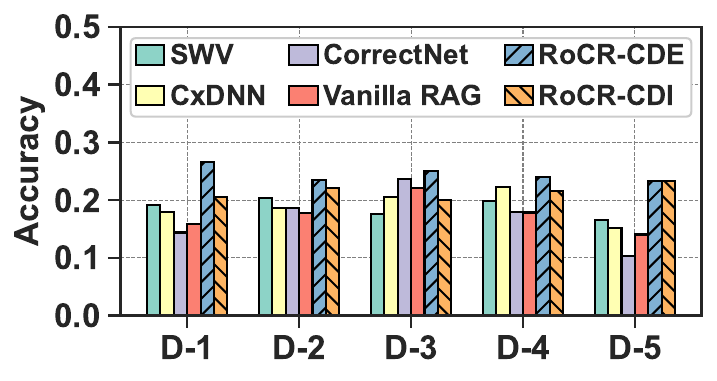}
    \caption{Movie on Phi-2}
    \label{fig:f}
  \end{subfigure}
    % \hfill % space between the subfigures
  \begin{subfigure}[b]{0.237\textwidth}
    \includegraphics[width=1\textwidth]{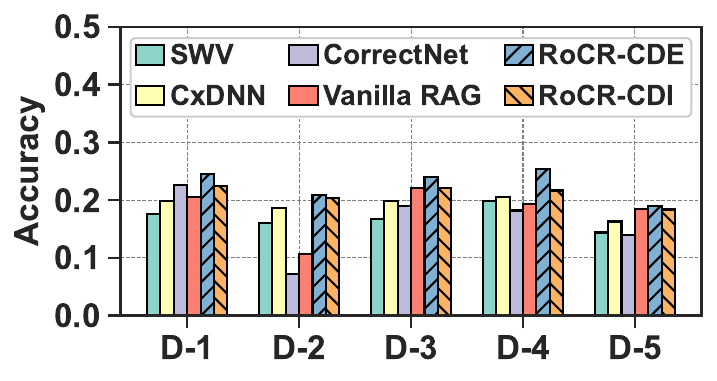}
    \caption{Movie on Mistral-7B}
    \label{fig:g}
  \end{subfigure}
  % \hfill % space between the subfigures
  \begin{subfigure}[b]{0.237\textwidth}
    \includegraphics[width=1\textwidth]{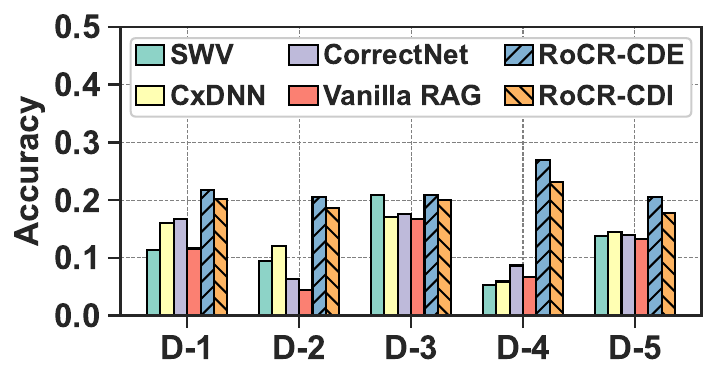}
    \caption{Movie on Llama-2-3B}
    \label{fig:h}
  \end{subfigure}
  \caption{Performance comparison between our framework and four baselines on RAG utilizing the LLMs including Gemma-2B, Phi-2, Mistral-7B, and Llama-2-3B with device variation  \ydel{\(\sigma = 0.1\)}\yadd{specified in Table~\ref{tab:var}}, given dataset \(Citation\) and \(Movie\). }
  \label{fig:four_LLMs}
\end{figure*}

\subsubsection{\textbf{Default Experimental Setting.}} 

Our framework chooses \textit{all-MiniLM-L6-v2} \cite{huggingface2023} as the sentence embedding model. 
% \textcolor{orange}{this sentence is not needed->} As shown in Figure ~\ref{fig:2}, the sentence embedding model converts the user query and past user data from user database from textual format to numerical representations. 
For each dataset, we randomly select 2000 documents from profile data as the anchor examples. To examine the data construction method of CDE, we set the augmentation factor \(k = 5\) to obtain 10000 negative and positive examples. 
% \textcolor{orange}{ and positive example pairs } examples \textcolor{orange}{not needed->} where one anchor example can generate five negative examples. 
We set dropout rate as 0.1 to obtain the positive examples while maintain it as 0 when process anchor and negative examples. To examine the data construction method CDI, we set dropout rate for positive examples as 0.1 and dropout rate for negative examples as 0.9. 
To align with experiments for CDE, we also set \(k = 5\) in the experiments for CDI.
% \textcolor{orange}{not needed, or just say "also using k=5 above"->}To align with the experiments for explicit labeled data, we also create five positive and five negative examples for each anchor example. 
For the experimental results, we run five times and get the average. 
% \textcolor{orange}{not needed->} The LLM, in Figure ~\ref{fig:2}, takes the MIPS-fetched data and the user query to generate response.
In experiments, we set the device variation \(\sigma = 0.1\) and shape embeddings into dimension of 64 with precision of \(int8\). The learning rate is \(2e-5\).

In all experiments, we adhere to the device variation model previously described. The specific parameters are abstracted and then simplified from three representative NVM devices, two of them are resistive random-access memory (RRAM) devices extracted from~\cite{yao2020fully, liu2023architecture} and the other is a ferroelectric field effect transistor (FeFET) device extracted from~\cite{wei2022switching}. We name them $RRAM_1$, $RRAM_4$ and $FeFET_2$, respectively. We also extrapolate the modeling data to obtain two synthesized $FeFET_3$ and $FeFET_6$ devices. Detailed device modeling results are demonstrated in Table~\ref{tab:var}. A $x$-level device means this device can represent $x$ distinct values \rqin{and $\sigma_{L_2} = 0.01$ means the variation of this device is 0.01 when it is representing the level value 2.} \rqin{Using the device variations obtained from real CiM devices, we perform our experiments on a single Nvidia A10 \yadd{GPU}.}

\begin{table}[t]
\footnotesize
    \centering
    \caption{
    Device non-ideality modeling for different real and synthesized devices. For devices with more than two levels, the device variation for each level is depicted as $L_x$. 
    }
    \begin{tabularx}{\columnwidth}{c*{5}{Y}}        
    \toprule
        \multirow{2}{*}{Name} & \multirow{2}{*}{\# of Levels}  & \multicolumn{4}{c}{Device Variations $\sigma_v$} \\
                  &   & $L_0$ & $L_1$ & $L_2$ & $L_3$ \\
        \midrule
        $RRAM_1$ (Device-1)  & 1 & 0.0100 & 0.0100 & 0.0100 & 0.0100\\
        $FeFET_2$ (Device-2) & 4 & 0.0067 & 0.0135 & 0.0135 & 0.0067\\
        $FeFET_3$ (Device-3) & 4 & 0.0049 & 0.0146 & 0.0146 & 0.0049\\
        $RRAM_4$ (Device-4)  & 4 & 0.0038 & 0.0151 & 0.0151 & 0.0038\\
        $FeFET_6$ (Device-5) & 4 & 0.0026 & 0.0155 & 0.0155 & 0.0026\\
        \bottomrule
    \end{tabularx}
    \label{tab:var}
\end{table}
% \vspace{0.1cm}

\rqin{Document embeddings are shaped based on different CiM devices and stored as parallel arrays, similar to how they would be mapped to multiple NVM devices in practical scenarios. For example, if an embedding is shaped to contain all uint8 values, when it is mapped to 4-level (2-bit) devices such as $FeFET_2$, each element of the vector is represented by four devices.
}
% Document embeddings are shaped based on different CiM desiges and then each element is mapped to multiple NVM devices. For example, if an embedding vector contains values quantized to unint8, when it's mapped to 4-level (2-bit) devices like $FeFET_2$, each element of this vector is represented by 4 devices.

\subsubsection{\textbf{Evaluation Methods.}}

Our first three datasets examine the model classification capability, and the rest of two datasets examine the text generation capability. In particular, dataset \(Citation\) and \(Movie\) has two and fifteen labels respectively. We can examine the binary and multiclass classification capabilities of the LLMs enhanced by our framework. In this way, we use accuracy to examine the ability of the models to correctly classify instances across different classes, and we use F1 score to examine the balance between precision and recall in classification tasks. For dataset \(Rating\), while it has five labels and also examine the multiclass classification, we use mean absolute error (MAE) and root mean square error (RMSE) to evaluate from from a regression perspective \cite{chai2014root}. For MAE, it measures the average magnitude of errors in the predictions, providing a straightforward assessment of the model's overall accuracy in predicting the rating values. For RMSE, it captures the square root of the average squared differences between predicted and actual ratings, offering a metric sensitive to larger errors, which can highlight significant discrepancies between the model's predictions and true values. For dataset \(News\) and \(DBLP\), their labels are sentences. Such datasets examine the text generation capabilities. We use ROUGE-1 and ROUGE-L to evaluate the overlap between generated texts and reference texts \cite{lin2004rouge}, capturing both the precision and recall of individual words (ROUGE-1) and the longest matching sequence (ROUGE-L), ensuring a comprehensive evaluation of the text generation quality. For accuracy, F1, ROUGE-1 and ROUGE-L, their higher values reflect the better performance. For MAE and RMSE, their lower value represent the better performance. 
Additionally, we use accuracy to measure the MIPS performance (MIPS accuracy), representing the ratio of MIPS results under device variation and MIPS results without device variation (references).

\subsubsection{\textbf{Baselines.}}

As this is the first work to improve the RAG robustness on Edge-based CiM, we do not have state-of-the-art for comparison. As such, we construct baselines from the past noise mitigation methods originally designed to boost DNN robustness. The first baseline is selective write verify \cite{yan2022swim} (\textbf{SWV}). While it originally utilizes the second derivation to evaluate the device variation impact on neural network weights, we use the second derivation to measure the embedding deviation between the ground truth embedding and the embedding under device variation. The second baseline is (\textbf{CxDNN}) \cite{jain2019cxdnn}. While they use compensation factor to improve the robustness of vector-matrix multiplication, we use the compensation factor the calibrate the embedding impacted by device variation. The third baseline is \textbf{CorrectNet} \cite{eldebiky2023correctnet}, where it utilizes the cross entropy loss and regularization to improve the robustness of neural networks in CiM. To use it as a baseline, we also use the cross entropy loss the regularization as the loss function to calibrate the device output embedding.
% While they utilize the cross entropy loss and regularization to improve the robustness of neural networks in CiM, \textcolor{orange}{take a look, not sure what the meaning is->} we use the cross entropy loss the regularization as the loss function to calibrate the device output embedding. 
Additionally, we examine the \textbf{Vanilla RAG}, which contains no noise mitigation methods, as our fourth baseline. The baselines use the same experimental setting as our framework does.

\subsection{Results}

% \textcolor{red}{We start with cmpare the MIPS accuracy [done]}

% \textcolor{red}{Then we present a comprehensive study of RAG, which contains MIPS and LLM, chose Gemma-2 [done]}

% \textcolor{red}{Additionally, choose two datasets, we examine four different LLMs. Why? We want to show that MIPS results can impact all different LLMs [done]}

% \textcolor{red}{Finally, we examine the impact of device variation}

For RAG, it can be simplified as the combination of MIPS and LLM, where the MIPS as a retriever searches the appropriate information and the LLM as a generator processes the searched results. Hence, in our experiments, we first evaluate the performance of MIPS under the device variation of device-1.
% We use MIPS accuracy as the metric
\rqin{We take the MIPS results obtained without device variation as the references \yadd{(\emph{i.e.}, ground truth)}. 
Using the metric of MIPS accuracy, we examine how many MIPS results under device variation will match the references. 
Since the quality of retrieved content largely depends on the base sentence embedding model, and we focus on mitigating the device variation impact on the embedding model, we do not assess the quality of references.}

As shown in Table ~\ref{tab:prelim_1}, our framework using the two data construction methods outperforms the four baselines across five datasets. 
% \textcolor{orange}{It is not that we can "restore the corruption", it is more like our method mitigates the corruption and is more robust.} 
It shows that our framework can mitigate the embedding \ydel{corruption}\yadd{perturbation} due to device variation.
% It shows that our framework can restore MIPS results from the corruption of device variation. 
These results can also correspond to the preliminary study shown in Figure ~\ref{fig:prelim}, \rqin{where the increment of $\sigma$ in naive Gaussian noise} will jeopardize the MIPS performance.

% \begin{table}[htbp]
% \caption{RAG performance using Gemma-2B on different CiM devices. Examine \(Baseline4\).}
%     \centering
%     \footnotesize
%     \begin{tabularx}{\columnwidth}{c*{5}{Y}} 
%         \toprule
%         \multirow{2}{*}{\textbf{Dataset}} & \textbf{Citation} & \textbf{Movie} & \textbf{Rating} & \textbf{News} & \textbf{DBLP} \\
%                     & Acc       & Acc       & MAE       & ROUGE-1   & ROUGE-1       \\
%         \midrule
%         Raw RAG     & 0.5200    & 0.3728    & 0.3150    & 0.0855    & 0.2295 \\
%         RAG Devi-1  & 0.4401    & 0.1017    & 0.3903    & 0.0754    & 0.1731 \\
%         RAG Devi-2  & 0.4801    & 0.1576    & 0.4153    & 0.0296    & 0.1618 \\
%         RAG Devi-3  & 0.4852    & 0.1614    & 0.3255    & 0.0725    & 0.1647 \\
%         RAG Devi-4  & 0.2101    & 0.1219    & 0.4015    & 0.0505    & 0.2219 \\
%         RAG Devi-5  & 0.4256    & 0.1152    & 0.3951    & 0.0676    & 0.2018 \\

%         \bottomrule
%     \end{tabularx}
%     \label{tab:prelim_1}
% \end{table}

\begin{table}[htbp]
\caption{Performance (MIPS accuracy) comparison between our framework and baselines. Accuracy is computed based on MIPS-retrieved documents under device variation of device-1 and the these retrieved without device variation.}
\vspace{-0.2cm}
    \centering
    \footnotesize
    \begin{tabularx}{\columnwidth}{c*{5}{Y}} 
        \toprule
        \textbf{Dataset} & \textbf{Citation} & \textbf{Movie} & \textbf{Rating} & \textbf{News} & \textbf{DBLP} \\
        \midrule
        SWV         & 0.4200    & 0.1728    & 0.1050    & 0.0855    & 0.2295 \\
        CxDNN       & 0.4401    & 0.2017    & 0.0503    & 0.0754    & 0.1681 \\
        CorrectNet  & 0.4013    & 0.0699    & 0.0509    & 0.0533    & 0.1609 \\
        Vanilla RAG & 0.4547    & 0.1694    & 0.0933    & 0.0649    & 0.1747 \\
        \textbf{RoCR-CDE}& \textbf{0.9231}& \textbf{0.4639}& \textbf{0.1583}& \textbf{0.1921}& \textbf{0.2750} \\
        \textbf{RoCR-CDI}& \textbf{0.9344}& \textbf{0.4355}& \textbf{0.1266}& \textbf{0.1708}& \textbf{0.2905} \\

        \bottomrule
    \end{tabularx}
    \label{tab:prelim_1}
\end{table}

% \textcolor{orange}{Is it possible to move the main table here?}
After we compare the MIPS performance of our framework and baselines, we further present a comprehensive evaluation to show the RAG performance of them. We use Gemma-2B as the LLM in RAG. Additionally, with Gemma-2B, we run RAG without device variation to obverse its \ydel{theoretical}\yadd{ideal} performance, where we get 0.5200 of accuracy for Citation, 0.3728 of accuracy for Movie, 0.3150 of MAE for Rating, 0.0855 of ROUGE-1 for News, and 0.2295 of ROUGE-1 for DBLP. On five CiM devices, whose device variations have been shown in Table ~\ref{tab:var}, we examine RAG with five datasets. 
% \textcolor{orange}{delete sentence->} As shown in Table ~\ref{tab:main}, our framework can mitigate the different device variation with respect to different datasets. 
As shown in Table ~\ref{tab:main}, given the same datasets, it is clear that each device variation significantly compromises the RAG robustness, whereas our framework can mitigate the different device variation.
% \textcolor{orange}{As shown in Table ~\ref{tab:main}, } given the same datasets, \textcolor{orange}{it is clear that each ...} we can observe that each device variation significantly compromises the RAG robustness. \textcolor{orange}{, whereas our framework can mitigate the different device variation.} 
For example, the RAG performance for Citation dataset on Device-2 can range from 0.18 to 0.48, while our framework can boost the accuracy performance of Citation dataset above 0.5 for all five devices. 
% \textcolor{orange}{move red to after "we use gemma 2b as the LLM...} \textcolor{red}{Additionally, using Gemma-2B, we run RAG without device variation to obverse its theoretical performance. We get 0.5200 of accuracy for Citation, 0.3728 of accuracy for Movie, 0.3150 of MAE for Rating, 0.0855 of ROUGE-1 for News, and 0.2295 of ROUGE-1 for DBLP.} 
Compared to the four baselines whose performances are relatively worse than the \ydel{theortical}\yadd{ideal} performance, our framework significantly approaches and sometimes outperforms the \ydel{theoretical}\yadd{ideal} performance via generating better sentence embeddings. \yadd{This is because RoCR also serves as a regularization to improve the model's generalization.}

In addition, we evaluate the impact of different LLMs on the performance of our framework. As Figure ~\ref{fig:2} shown, the LLM takes the concatenation of MIPS searched data and user query as the input and generates the response regarding the user query. Since different LLMs may have different response given the same query, we select four emerging edge-friendly medium-size LLMs in our experiments to examine the performance of our framework. Gemma-2B \cite{team2024gemma} is a new SOTA open model introduced by Google, with 4.95G model weights. According to Google, Gemma can outperform the same sized Llama-2 in reasoning capabilities. Hence, we also use Llama-2-3B \cite{openlm2023openllama}, one of the earliest open LLMs introduced by Meta, with 6.85G model weights. Similarly, Phi-2 \cite{gunasekar2023textbooks} released by Microsoft, is a powerful small LLM with 5G model weights. Additionally, Mistral-7B-GPTQ \cite{jiang2023mistral} made by Mistral AI, is a well-performed LLM after Llama model. We select dataset \(Citation\) and dataset \(Moive\). We use the default experimental setting with \(\sigma\) = 0.1 and use CiM Device-1 as the experimental environment. The results are shown on Figure ~\ref{fig:four_LLMs}. It is evident that our framework outperforms each baseline across five CiM devices. Besides, the performance of each baseline on the same dataset can be largely different given different device, while our framework can produce a more robust performance. 

\begin{figure}[h]
  \centering
  \includegraphics[width=0.75\columnwidth]{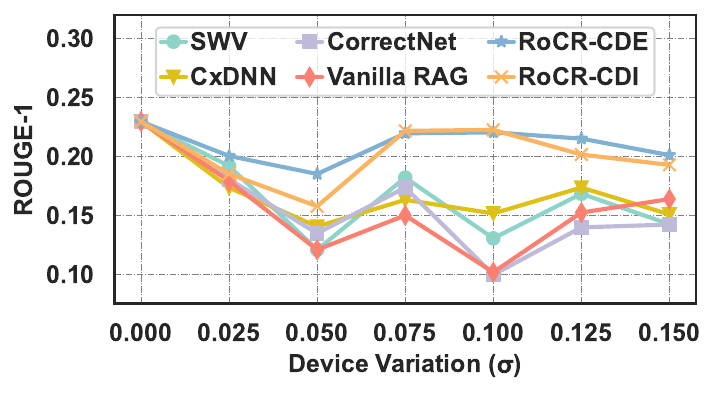}
  % \vspace{-0.4cm}
  \caption{Performance comparison between our framework and four baselines on CiM device-1 with different device variation \(\sigma\), given dataset DBLP.}
  \label{fig:ablation1}
  \vspace{-0.2cm}
\end{figure}

By default, we use \(\sigma = 0.1\) to calculate the device variation of the five CiM devices. We also conduct an additional study to evaluate our framework given different \(\sigma\) values. Since we have already use dataset Citation and dataset Movie to study the performance of our frameworks seen in Figure ~\ref{fig:four_LLMs}, we choose a different dataset DBLP, using ROUGE-1 as the metric. For the LLM in RAG, we choose Mistral-7B. We examine the \(\sigma\) values higher and lower than 0.1, including 0, 0.025, 0.05, 0.075, 0.125, and 0.15. \rqin{The case of \(\sigma\) = 0 reflects the ideal performance}.  For the CiM device, we use CiM device-1. As shown in Figure ~\ref{fig:ablation1}, our framework outperforms baselines across different device variation values.

\ydel{\rqin{Finally, when our framework is utilized on CiM-backed RAG, it only makes inference and generates the noise-resilient embeddings. In this way, it will not cause additional overhead.
}}
\yadd{
Finally, RoCR is a training method that generates more robust weights for the sentence embedding model. It does not change the model structure. Thus, there is no hardware (\emph{e.g.}, energy and latency) overhead during inference.
}
% \textcolor{red}{make line more readable, use rouge}

% In Figure ~\ref{fig:overhead}, we study the overhead. \textcolor{red}{[the figure is a sample, we will get results soon...]}

% \begin{figure}[h]
%   \centering
%   \includegraphics[width=0.7\columnwidth]{figures/fig_6.png}
%   \caption{overhead for each method}
%   \label{fig:overhead}
% \end{figure}

% \textcolor{red}{here I am going to add more details including explanation and analysis. But I will add some other experimental chart first}

\section{Conclusion}

In this paper, we present a novel framework for retrieval-augmented generation (RAG) acceleration via computing-in-memory (CiM) architectures. Our approach provide a solution to free RAG from the constraints of latency and scalability on edge devices. By optimizing the sentence embedding model, our framework enable the utilization of CiM devices in storing and processing the document embeddings, minimizing the impact of CiM device variations. Experimental results show that our framework achieves superior RAG performance and largely mitigates the impact of device variations. This paper marks the first RAG acceleration via CiM framework.

% \clearpage % Ensures bibliography starts on a new page
\bibliographystyle{unsrt}
\bibliography{citations}
\balance

\end{document}